\documentclass[preprint,12pt,nonatbib]{elsarticle}

\usepackage{amssymb}
\usepackage{subcaption} 
\usepackage{multirow}
\usepackage{apacite}

\journal{Expert Systems with Applications}

\begin{document}

\begin{frontmatter}

Leveraging Explainable AI for LLM Text Attribution: Differentiating Human-Written and Multiple LLMs-Generated Text
\\
\\
Authors’ names and affiliations: 
\begin{itemize}	
\item Ayat Najjar is with Arab American University, Jenin P.O Box 240, Palestine. e-mail: ayat.najar@aaup.edu.
\item Huthaifa I. Ashqar is with Arab American University, Jenin P.O Box 240, Palestine and Columbia University, New York, NY 10027, USA. e-mail: huthaifa.ashqar@aaup.edu.
\item Omar Darwish is with Eastern Michigan University, Ypsilanti, MI 48197, USA. e-mail: odarwish@emich.edu.
\item Eman Hammad is with iSTAR Lab at Texas A\&M University, College Station, TX 77840, USA. e-mail: eman.hammad@tamu.edu.
\end{itemize}

Address for corresponding author:
Eman Hammad, Office 301, Fermier Hall, 106 Ross St. Texas A\&M University, College Station, TX 77843-3367. USA. Email: eman.hammad@tamu.edu.

\newpage

\title{Leveraging Explainable AI for LLM Text Attribution: Differentiating Human-Written and Multiple LLMs-Generated Text}






\begin{abstract}
The development of Generative AI Large Language Models (LLMs) raised the alarm regarding identifying content produced through generative AI or humans. In one case, issues arise when students heavily rely on such tools in a manner that can affect the development of their writing or coding skills. Other issues of plagiarism also apply. This study aims to support efforts to detect and identify textual content generated using LLM tools. We hypothesize that LLMs-generated text is detectable by machine learning (ML), and investigate ML models that can recognize and differentiate texts generated by multiple LLMs tools. We leverage several ML and Deep Learning (DL) algorithms such as Random Forest (RF), and Recurrent Neural Networks (RNN), and utilized Explainable Artificial Intelligence (XAI) to understand the important features in attribution. Our method is divided into 1) binary classification to differentiate between human-written and AI-text, and 2) multi classification, to differentiate between human-written text and the text generated by the five different LLM tools (ChatGPT, LLaMA, Google Bard, Claude, and Perplexity). Results show high accuracy in the multi and binary classification. Our model outperformed GPTZero with 98.5\% accuracy to 78.3\%. Notably, GPTZero was unable to recognize about 4.2\% of the observations, but our model was able to recognize the complete test dataset. XAI results showed that understanding feature importance across different classes enables detailed author/source profiles. Further, aiding in attribution and supporting plagiarism detection by highlighting unique stylistic and structural elements ensuring robust content originality verification.
\end{abstract}


\begin{highlights}
\item Curating a text attribution dataset containing human-written and LLM generated texts
\item Leveraging ML and XAI to attribute text generated by five LLM tools and humans
\item A path towards textual content attribution in human and multi-LLM generative AI world
\end{highlights}


\end{frontmatter}


\section{Introduction}

The recent development of generative AI models is rapidly altering our communication patterns. It is extensively utilized in many fields, including the arts, healthcare, education, and content creation. The subject at hand is the growing risks associated with plagiarism in academic contexts, especially as more and more students turn to LLM tools like ChatGPT, Google Bard, Claude, etc. to produce academic writing~\cite{hadi2023survey}. The unethical taking of another person's work without giving due credit is known as plagiarism, and it has long been a problem in educational institutions~\cite{awasthi2019plagiarism}. The introduction of sophisticated language models exacerbates this issue. One major concern is that students' writing skills may deteriorate because of their heavy reliance on AI tools for assignments and essays, which will impede the crucial process of developing their expressive capabilities.

In addition to compromising the integrity of the educational system, this reliance on AI-generated content makes it difficult for teachers to evaluate students' academic progress and true talents fairly. Therefore, the pressing problem requires a thorough investigation and a calculated response to deal with the complex effects of unrestricted plagiarism made possible by sophisticated language models~\cite{tami,sam}.

The importance of this paper lies in its contribution to the evolving field of textual content attribution in a world where both human-authored and AI-generated text, especially from multiple large language models (LLMs), are increasingly indistinguishable. As AI systems like ChatGPT, LLaMA, Google Bard, Claude, and Perplexity~\cite{chatgbt, llama2, bard, claude, pplx} become more prevalent, ensuring the authenticity and trustworthiness of content has become a major concern in various domains, such as cybersecurity, academic integrity, and business operations~\cite{rad,mas}.

This paper addresses a critical need to reliably differentiate between human-written and AI-generated content, which has significant implications for maintaining ethical standards, ensuring information security, and fostering accountability in an AI-driven world. By proposing an effective model that outperforms generalized systems like GPTZero, this research offers a solution to the growing challenge of content attribution, paving the way for more accurate detection and classification of text originating from diverse sources—whether human or multiple AI systems.

In a world where multiple LLMs are generating content for varied purposes~\cite{jar1,kim}, from automated reports to news articles, the ability to trace the origin of the text is crucial~\cite{jar3,nam}. This study’s approach of utilizing machine learning algorithms and Explainable AI (XAI) provides not only high accuracy but also transparency in classification, making it valuable for ensuring the integrity of digital content across industries. As AI continues to shape communication and content creation, the findings of this paper will play a vital role in shaping how we verify and attribute textual content in a multi-LLM, generative AI environment.

The rest of this study is structured as follows: In Section II, we offer a thorough analysis of relevant research in the area of LLMs-generated text identification using ML. Our suggested strategy is thoroughly explained in Section III. We assess our approach's performance and provide the findings in Section V. Finally, we wrap up our paper in Section VI.

%
%
%
%

 

\section{Related Works}
\label{sec:relatedwork}

The study seeks to address the growing challenge of textual content attribution in a world where AI systems, such as LLMs, are increasingly used to generate content. By testing and comparing ML and DL algorithms, the study aims to demonstrate how a fine-tuned, targeted approach can outperform more generalized models in accurately classifying text and ensuring transparency in the decision-making process through XAI techniques. Specifically, we will look at two sections: one for multi-classification, in which we will distinguish between text written by humans and five different LLM tools (ChatGPT, LLaMA, Google Bard, Perplexity, and Claude); and another for binary classification, in which we will distinguish between text generated by LLMs generally and text written by humans.

\subsection{Advancements in Text Attribution} 
These studies explore creative approaches to text attribution, exposing a dynamic evolution in methods that combine linguistic analysis, deep learning, and machine learning. The various methods highlight a dedication to improving accuracy and flexibility in the area of plagiarism detection. 

Changing the words or word order in a statement to create a different version is known as paraphrasing or rephrasing. In NLP, identifying paraphrases is a difficult problem~\cite{jar2}. By using the SVM, logistic regression, and RNN algorithm models, this study~\cite{hunt2019machine} seeks to identify instances of paraphrase plagiarism, the best method that provides accuracy with 80\% is RNN. Using four well-known models—Bag of Words (BOW), Latent Semantic Analysis (LSA), Support Vector Machine (SVM), and Stylometry this study~\cite{alsallal2016integrated} aims to provide a unified method for plagiarism detection. Using 25 books by different writers, the study analyzes data based on how frequently the Most Common Words (MCW) are used. The increased weighting approach of the adjusted LSA performs better than the conventional LSA method, according to the results. An additional study~\cite{anguita2011automatic} presents a new approach to detect cross-language plagiarism by machine learning and natural language processing. The procedure is as follows: textual input, translation detection, online search, and report production. Most documents with electronic input can be used with this method. Findings demonstrate that the system can locate instances of Spanish materials that have been plagiarized online from English sources, both by humans and by machines. In 56\% of the cases, the system was able to identify the source of plagiarism, this proportion rises to 67\% in the case of machine translation. With the primary goal of identifying plagiarism in source codes in mind, this study~\cite{kikuchi2014source}  suggests a plagiarism detector that is insensitive to variations in program statement order or identifiers. It compares its methodology with simulation-based plagiarism detection, integrating many Syntax tree components and Sequence Alignment into the system. Moreover, they disclose how their approach effectively identifies instances of plagiarism. Another study~\cite{suleiman2017deep} suggests utilizing the word2vec model, which is a model for detecting plagiarism in Arabic literature using Deep Learning characteristics. This approach evaluates the semantic similarity between Arabic words by using cosine similarity, which provides a highly accurate way to compare vector similarity. The similarity measures illustrate how even minor textual modifications, like swapping out a word or shifting the order of verbs and nouns, can produce results with a similarity value of 99\%, making it possible to identify plagiarism even in cases where test administration modifies the wording or substitutes synonyms for test items. 

\subsection{Leveraging LLMs for Text Detection} 
The use of LLMs for text detection has been the subject of the following studies, which highlight how they can improve textual content recognition accuracy and contextual understanding. This section examines relevant papers that explore the use of LLMs to improve text detection capabilities. 

A review of the existing LLM-generated text detection methods is provided by two surveys~\cite{tang2024science, wu2023survey}. The first survey~\cite{tang2024science} The purpose of this survey is to improve language generation model control and regulation while offering a summary of current LLM-generated text detection methods. In addition, we highlight important directions for future work to advance LLM-generated text detection, such as the creation of thorough assessment criteria and the danger of open-source LLMs. The second survey~\cite{wu2023survey}  gathers the most recent findings in this field of study and emphasizes the urgent necessity to support detector research in this survey. It also delves into widely used datasets, explaining their shortcomings and future development needs. Moreover, it examines different LLM-generated text recognition paradigms, illuminating issues such as data ambiguity, possible assaults, and out-of-distribution issues. In summary, it indicates promising avenues for further investigation into LLM-generated text detection to progress the application of responsible artificial intelligence (AI). With this study, we hope to give novices a thorough introduction to the topic of LLM-generated text identification, as well as seasoned researchers a useful update. 

In order to conduct a comparison analysis, a novel dataset comprising human-written and LLM-generated texts throughout the many genres included in this study~\cite{hayawi2023imitation} such as stories, poems, essays, and Python code—is introduced. Their results demonstrate how well various machine learning models can differentiate between text created by AI and human input, with the best results being observed in binary classification tasks. However, there are problems with categorizing GPT-generated text, especially in narrative composition, which highlights the intricacy of multiclass assignments involving many LLMs. The dataset provides a basis for further research in this rapidly evolving subject, and their insights have significant implications for AI text identification. An additional study~\cite{orenstrakh2023detecting} initiative is to educate the public on the effectiveness of LLM-generated text detectors and their utilization in upholding academic integrity. According to their findings, GPTKit is best for minimizing false positives, GLTR is the most robust, and Copy Leaks is the most accurate detector. On the other hand, GPTZero's false positives raise certain issues. The study highlights the detectors' shortcomings about code, non-English languages, and paraphrased information, highlighting the want for continual advancements to offer a complete remedy for maintaining academic integrity. We also propose ways to improve detector usability: simplifying API integration; providing clear documentation; and supporting widely used languages. Finally, this study~\cite{chen2023data} investigates a novel LLM-based strategy for data race detection that combines fine-tuning and motivating engineering methods. After DataRaceBench was used to create the specific DRB-ML dataset, fine-grain labels describing data race pairs, related variables, line numbers, and read/write information were added. By assessing exemplar LLMs and optimizing publicly available ones with DRB-ML, their study highlights the practicality of LLMs in data race identification. But they are not as effective as more conventional methods, especially when it comes to giving specific details on variable pairs that result in data races. 

\section{Methodology}
\label{sec:methodology}

\begin{figure}[t]
\begin{center}
\includegraphics[width = \columnwidth]{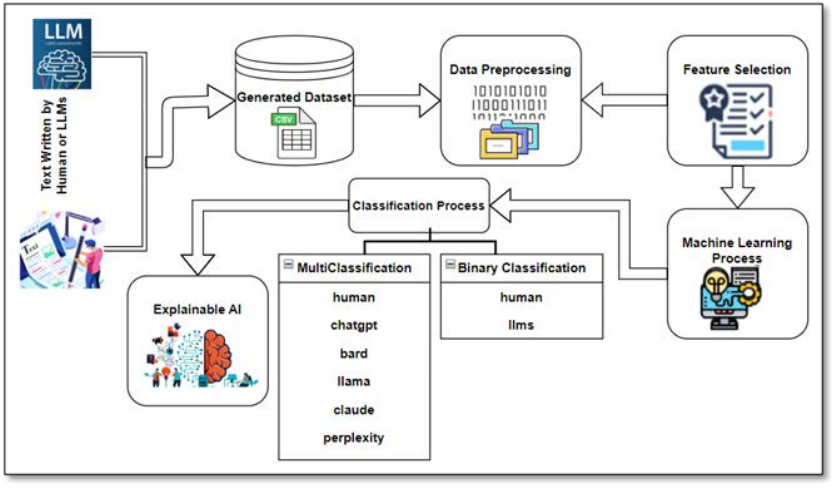}
\end{center}
\caption{The general workflow of the proposed ML model for detecting and classifying texts written by LLM}
\label{F1}
\end{figure}

Generally, data preparation and feature selection processes from a generated dataset play important roles in simplifying the subsequent tasks, like the classification task, leading to improved classification rates. This study proposes a framework for detecting the texts written by LLMs, shown in Figure~\ref{F1}, which includes four main phases, including data preprocessing, feature selection, machine learning model, and detection and classification. The following sections describe each step of this framework.


\subsection{Dataset Collation and Generation}

The dataset has 600 observations and was compiled in November 2023. It has 300 observations written by humans that are extracted from the Kaggle dataset for detecting texts written by LLMs~\cite{kaggle} this dataset has.  

\begin{itemize}
    \item id - A unique identifier for each essay. 

    \item prompt\_id - identifies the prompt to which the essay was produced. Two essays are available: "Car-free cities," rated “0”; and "Does the electoral college work?" rated “1”. 

    \item text - The essay text itself. 

    \item generated - Whether the essay was written by an LLM “1” or by a student “0”. 
\end{itemize}
    
and another 300 were produced manually by the author from 5 different LLMs (ChatGPT, LLaMA, Google Bard, Claude, and Perplexity) by asking each LLM to generate 30 essays for each subject of the essays mentioned before. There are three text-based attributes that each entry has: "text," "category," and "subcategory". By bridging the gaps between the academic and corporate worlds, this research project significantly impacts the fields of LLMs and plagiarism detection while also advancing the field of NLP. 

\subsection{Data Preprocessing and Feature Selection}

\begin{figure}[t]
\begin{center}
\subfloat[]{\includegraphics[width = 1.8 in]{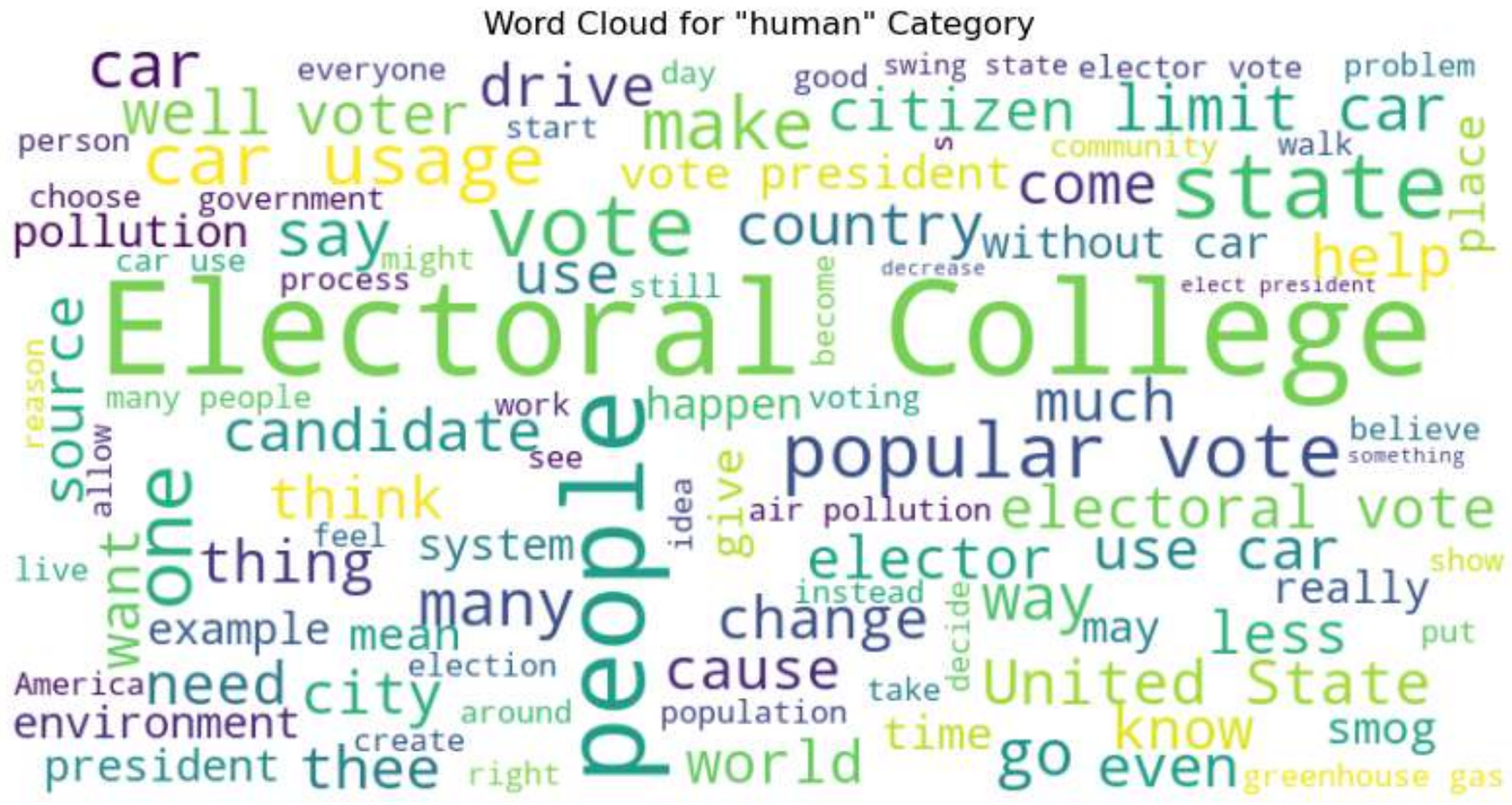}}
\subfloat[]{\includegraphics[width = 1.8 in]{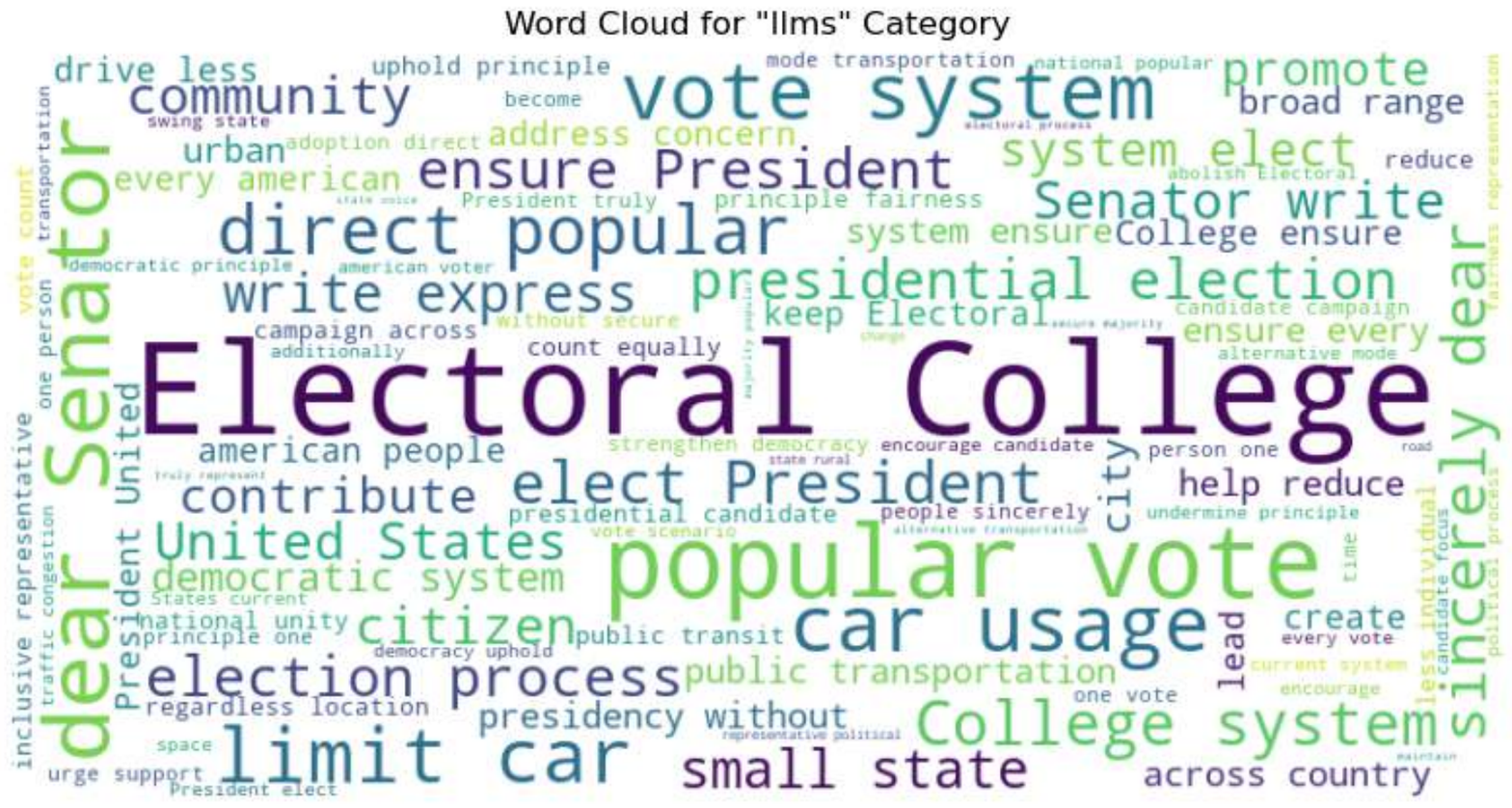}}            
\end{center}
\caption{Word cloud for (a)‘human’ and (b)‘LLMs‘classes}
\label{F2}
\end{figure}

A preliminary check was conducted during the Data Preprocessing phase to make sure no empty observations were detected. The text data was improved using common preparation techniques for NLP jobs. This featured stop words removal, lemmatization, punctuation removal, and tokenization of text~\cite{tabassum2020survey}. Putting the text data into a clean, structured format suited for classification and other NLP tasks helped prepare the dataset for later analysis and model creation. The word cloud for the two classes "human" and "llms" is displayed in Figure~\ref{F2}, and Table~\ref{tab1} displays the word frequency for the two classes as counts and percentages. this graphic and table following the use of the preprocessing step. After the data pretreatment stages, the following phase involved converting the category classifications into numerical representations. The use of machine learning algorithms that require numerical inputs was made possible by this change. The "category" feature, which distinguished between "human" and "llms" content, was specifically encoded into numerical labels, with the "human" class being represented by 0 and the "llms" class being represented by 1.



\begin{table}[]
\caption{Top 10 words for 'human' and ‘LLMs’ classes (counts and percentage of total tokens)}
\label{tab1}
\begin{tabular}{llllll}
\hline
\multicolumn{3}{l}{Words Frequency – Human Texts} & \multicolumn{3}{l}{Words Frequency – LLMs Texts} \\ \hline
Word           & Counts      & Percentage \%      & Word           & Counts      & Percentage \%     \\ \hline
car            & 2464        & 2.72               & system         & 337         & 2.28              \\ \hline
vote           & 2163        & 2.39               & Electoral      & 319         & 2.15              \\ \hline
people         & 1360        & 1.50               & vote           & 317         & 2.14              \\ \hline
state          & 1121        & 1.24               & college        & 303         & 2.05              \\ \hline
Electoral      & 958         & 1.06               & state          & 225         & 1.52              \\ \hline
would          & 901         & 0.99               & popular        & 182         & 1.23              \\ \hline
college        & 884         & 0.97               & car            & 178         & 1.20              \\ \hline
not            & 767         & 0.85               & ensure         & 177         & 1.20              \\ \hline
electoral      & 750         & 0.83               & would          & 161         & 1.09              \\ \hline
use            & 656         & 0.72               & dear           & 150         & 1.01             
\\ \hline
\end{tabular}
\end{table}

\begin{figure}[t]
\begin{center}
\subfloat[]{\includegraphics[width = 1.8 in]{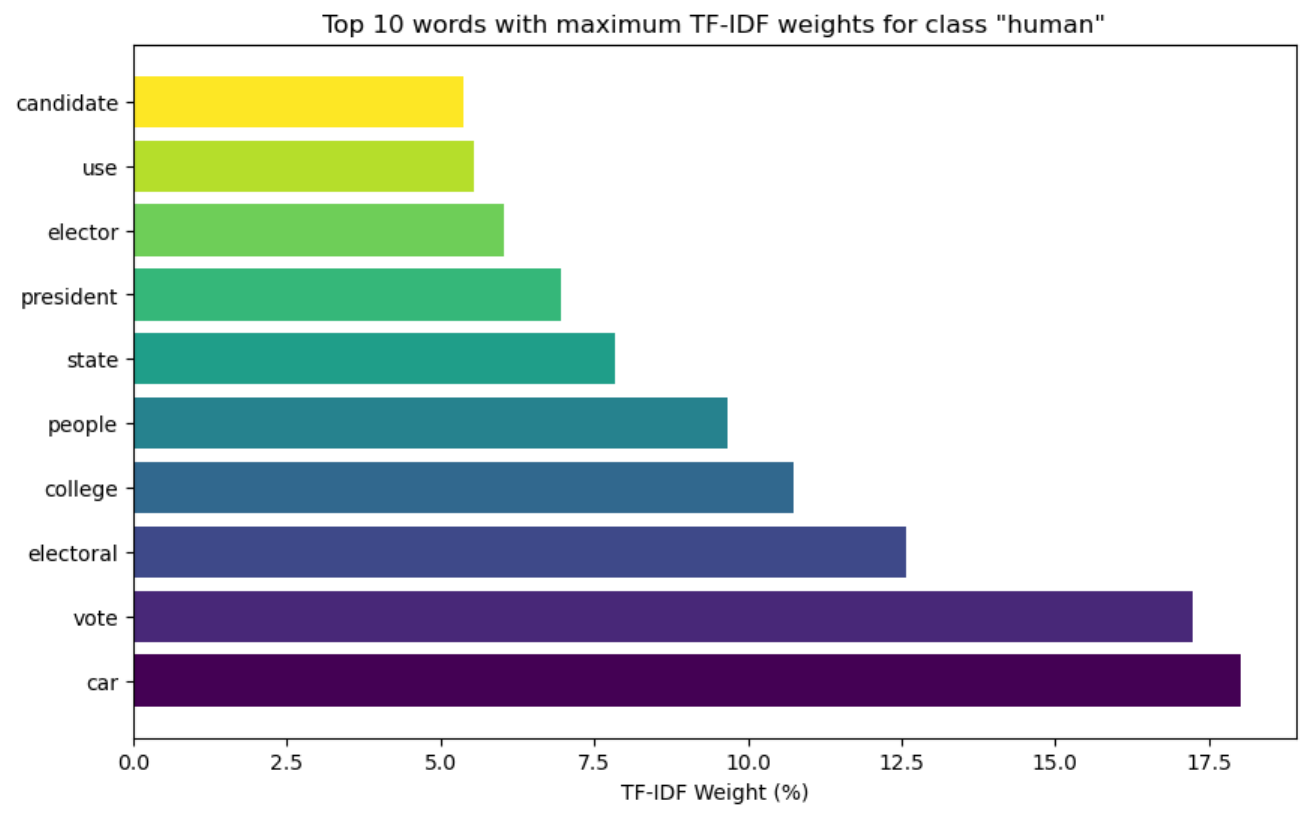}}
\subfloat[]{\includegraphics[width = 1.8 in]{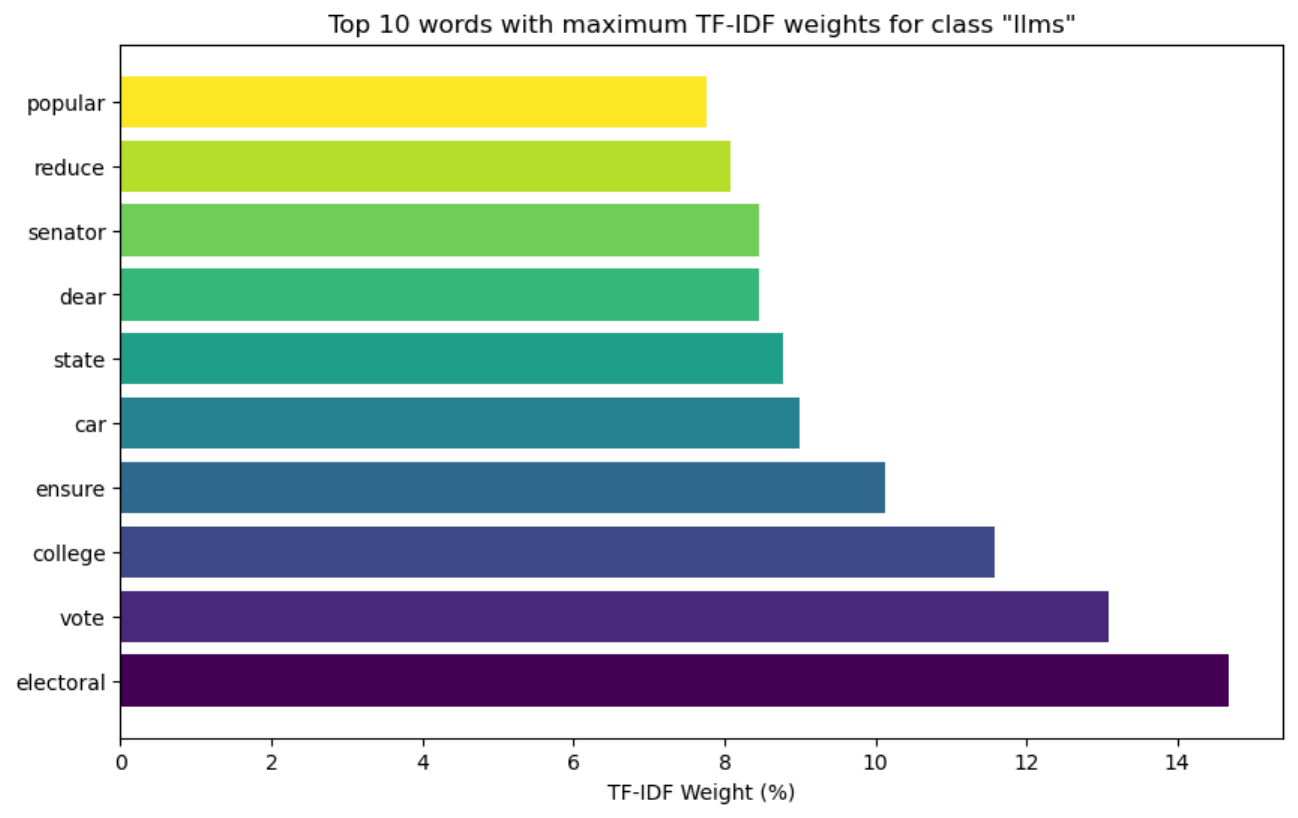}}   
\end{center}
\caption{Top 10 words with maximum TF-IDF weights for (a)‘human’ and (b)‘LLMs’ classes}
\label{F3}
\end{figure}

The dataset was then separated into training and testing subsets using an 80/20 split, with 80\% of the data going toward training and 20\% for testing. This division played a critical role in the model evaluation process by evaluating the model's performance on unseen data. Finally, a TF-IDF (Term Frequency-Inverse Document Frequency) Vectorizer was used to make it easier to convert the text data into a machine-learning-friendly format~\cite{abubakar2022sentiment}. By transforming the text data into a matrix of numerical features, this approach was able to capture the significance of words inside each document while considering their frequency across the entire dataset. With "0" denoting the “human” class and "1" denoting the “llms” class, the resulting TF-IDF vectors served as the basis for training machine learning models on this dataset, allowing the creation of classifiers to differentiate between human and “llms” generated texts. Figure~\ref{F3} displays the top 10 words with the highest TF-IDF weights for the two classes, "human" and "llms".


\subsection{Classification Algorithms} 

In the field of ML and classification, training models to produce predictions and classify texts authored by LLMs is an interesting and significant task. This technology allows computers to identify patterns in data and perform actions based on those patterns by using the RF and XGBoost algorithms [21] and RNN [22]. The precise identification of relevant text was made easier by this technique. 

RF: robust ensemble learning method, it is well-known for its performance in both classification and regression applications. With the help of a group of decision trees, it performs exceptionally well at managing complex datasets and reducing overfitting. Random Forest is a highly favored option in numerous fields due to its adaptability and resilience, yielding precise and dependable forecasts as well as valuable insights via feature importance analysis. The study also explored the rapidly changing neural network landscape, utilizing the power of (RNN) with a focus on texts generated by LLM’s analysis [23]. These cutting-edge deep-learning approaches produced ground-breaking capabilities, automating the complex pattern extraction LLM’s texts.  

XGBoost: Extreme Gradient Boosting, or XGBoost, is a ML technique that is notable for its great efficiency and scalability. The XGBoost algorithm, which is part of the gradient boosting family, performs exceptionally well and accurately in predictive modeling applications. XGBoost is a preferred option in many industries, including finance and healthcare, due to its capacity to manage complicated relationships in data, regularization techniques, and parallel processing. Acknowledged for its swiftness and efficiency, XGBoost has established itself as a mainstay in both practical and competitive ML scenarios. 

RNN: neural networks specifically constructed for sequential data processing. Because they have internal memory, they can retain knowledge about earlier inputs, which makes them appropriate for tasks like time series analysis and natural language processing. Even though they are good at capturing temporal dependencies, classic RNNs have problems with things like vanishing gradients. To overcome these constraints, more sophisticated topologies have been developed, such as Gated Recurrent Units (GRUs) and Long Short-Term Memory (LSTM) networks, which improve RNNs' ability to represent and learn intricate sequential patterns. 

Our text detection framework employs a two-stage classification process. Initially, binary classification distinguishes between two classes. Subsequently, multi-classification refines the identification by categorizing text into five classes, ensuring a comprehensive and nuanced approach to text detection. 

\subsubsection{Binary Classification} 

We added a column named “category” to the dataset that indicates the two classes: "human" to indicate that the text is written by a human and "llms" to indicate that the text is generated by one of the LLMs that we used (ChatGPT, LLaMA, Google Bard, Claude, and Perplexity).  

\subsubsection{Multi Classification} 

We added a column named “subcategory” to the dataset that indicates the two classes: "human" to indicate that the text is written by a human and another four classes "chatgpt", “LLaMA”, “Google Bard”, “Claude” or “Perplexity” to indicate that the text is generated by one of the LLMs that mentioned.  

\subsection{Explainable Artificial Intelligence (XAI)}

In recent years, artificial intelligence has advanced significantly, sparking interest in previously understudied fields. The focus has shifted from solely focusing on model performance as AI advances to requiring experts to look at algorithmic decision-making processes and the logic behind AI models' output. As modern ML algorithms especially "deep learning" ones using black box techniques become more powerful and complex, making it difficult to understand how they behave and why specific outcomes were achieved or mistakes were made, explainable artificial intelligence (XAI) systems are becoming more and more necessary. However, understanding those models' behaviors is equally as crucial as their outputs, allowing users to develop the proper level of trust and reliance [24], [25]. 

In the field of XAI, Local Interpretable Model-agnostic Explanations (LIME) is a crucial instrument that provides a way to understand how sophisticated ML models make decisions. Because LIME is based on a model-agnostic premise, which was developed by Ribeiro et al. in 2016 [26], [27], it can offer visible and interpretable insights into the predictions of different black-box models. LIME generates locally faithful approximations through perturbed samples around individual instances, enabling users to understand the reasoning behind individual predictions. Its interpretability-enhancing capabilities and adaptability have led to LIME's widespread adoption in various domains, where it is a valuable resource for researchers and practitioners seeking transparency in the decision-making process of complex ML algorithms. 

\section{Experimental Results and Discussion}
\label{sec:results}
This section assesses and examines the performance of different ML algorithms. An 11th generation Intel(R) Core (TM) i5-1135G7 @ 2.40GHz processor, 16.0 GB of RAM, and a 64-bit operating system were used in the experiment, as well as a Jupyter notebook was utilized to program in Python.  

In the first stage of our work, we focused on using the previously mentioned algorithms to distinguish the text written by LLMs vs the text written by humans which is binary classification. The accuracy results, for the three algorithms ranged from 94\% to 98\% as shown in Table~\ref{tab2}. Next, we decided to distinguish between the text written by humans and the text generated by which is multi-classification. The accuracy results ranged from 71\% to 97\% and are shown in Table~\ref{tab3}.  

The results demonstrate that, when it comes to binary classification, the accuracy of distinguishing between text written by humans and LLMs is very high. because literature created by humans exhibits a profound awareness of context, drawing from individual experiences and cultural quirks, and demonstrating creativity, emotional intelligence, and moral judgment. LLMs, on the other hand, produce replies based on learned patterns from enormous datasets and lack actual comprehension~\cite{bender2019data}. However, when multi-classification was discussed, it was noted that because there are already four classes from LLMs that might display similar characteristics in text composition, the accuracy tends to be lower than in binary classification. 

\subsection{Models Evaluation}

\subsubsection{Binary Classification }

In the binary classification, we used three different ML and NN techniques (RF, XGBoost, and RNN) as shown in Table~\ref{tab2}. Using RF, XGBoost, and RNN, we obtained an accuracy ranged between 94\% to 98\% and precisiona and recall of about the same range. This shows that the three algorithms correctly detected most of the observations and successfully distinguished between text that is written by humans and text that LLMS generate. 

\subsubsection{Multi-classification} 

\begin{figure}[t]
\begin{center}
\subfloat[]{\includegraphics[width = 1.2 in]{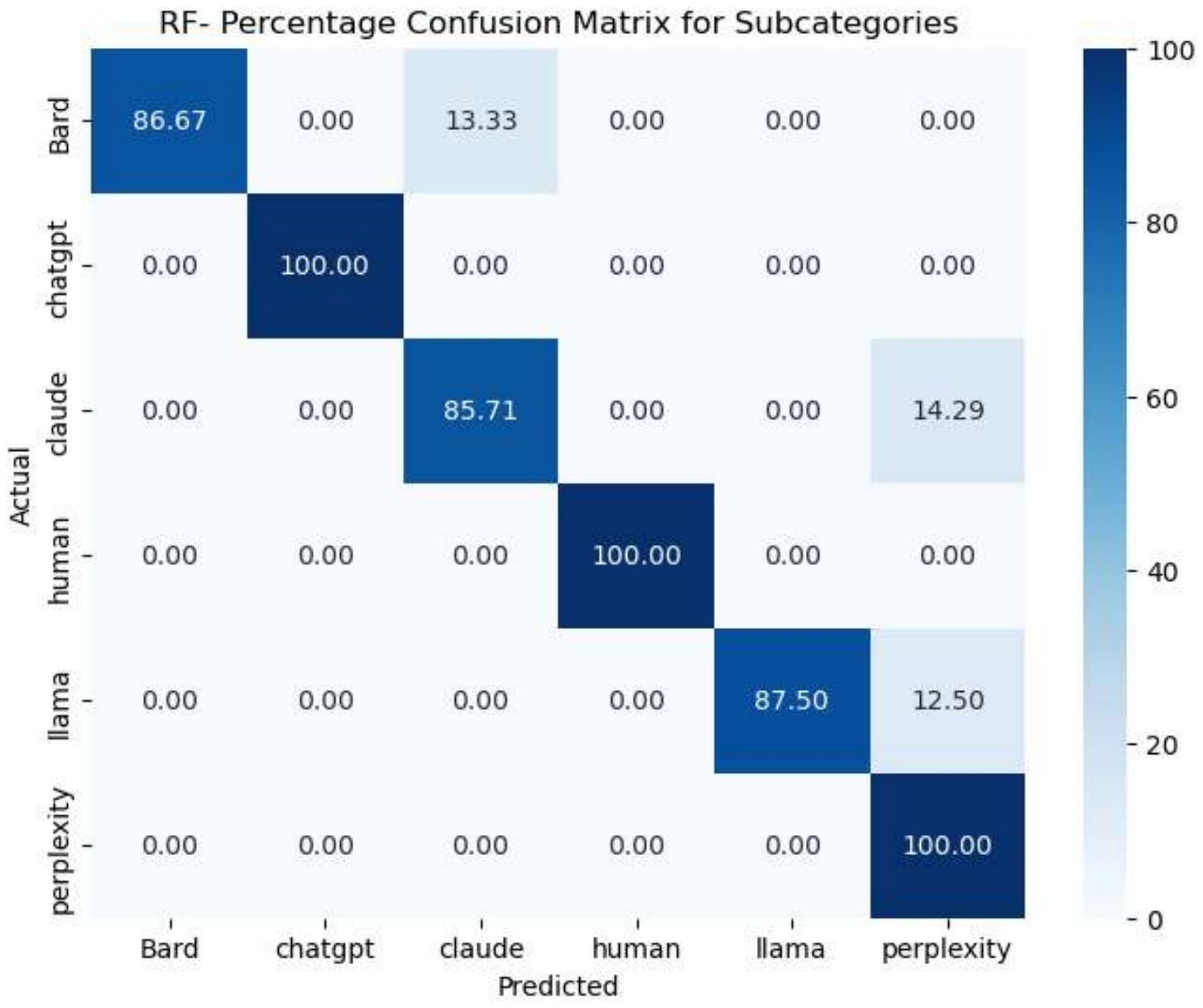}}
\subfloat[]{\includegraphics[width = 1.2 in]{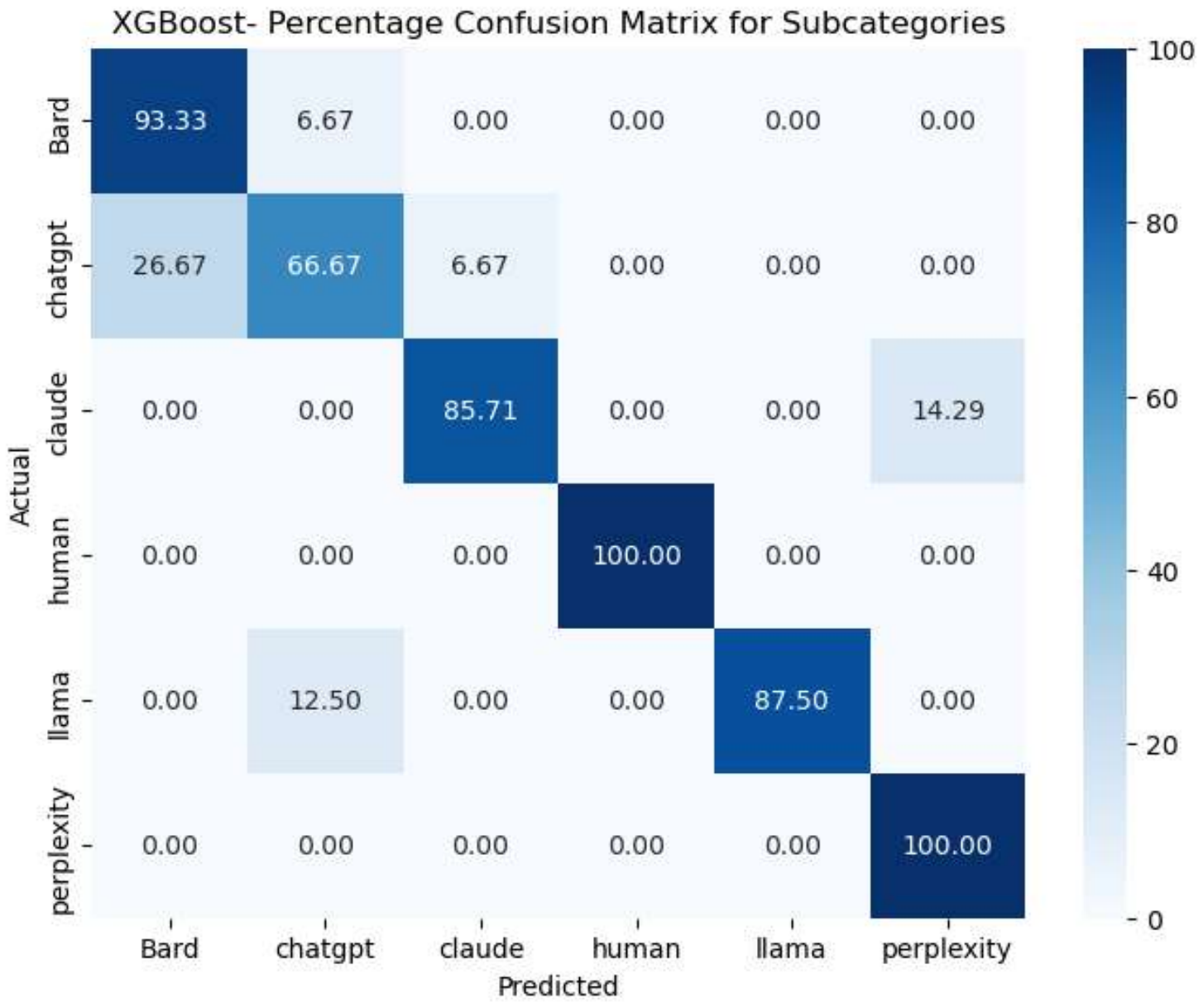}} 
\subfloat[]{\includegraphics[width = 1.2 in]{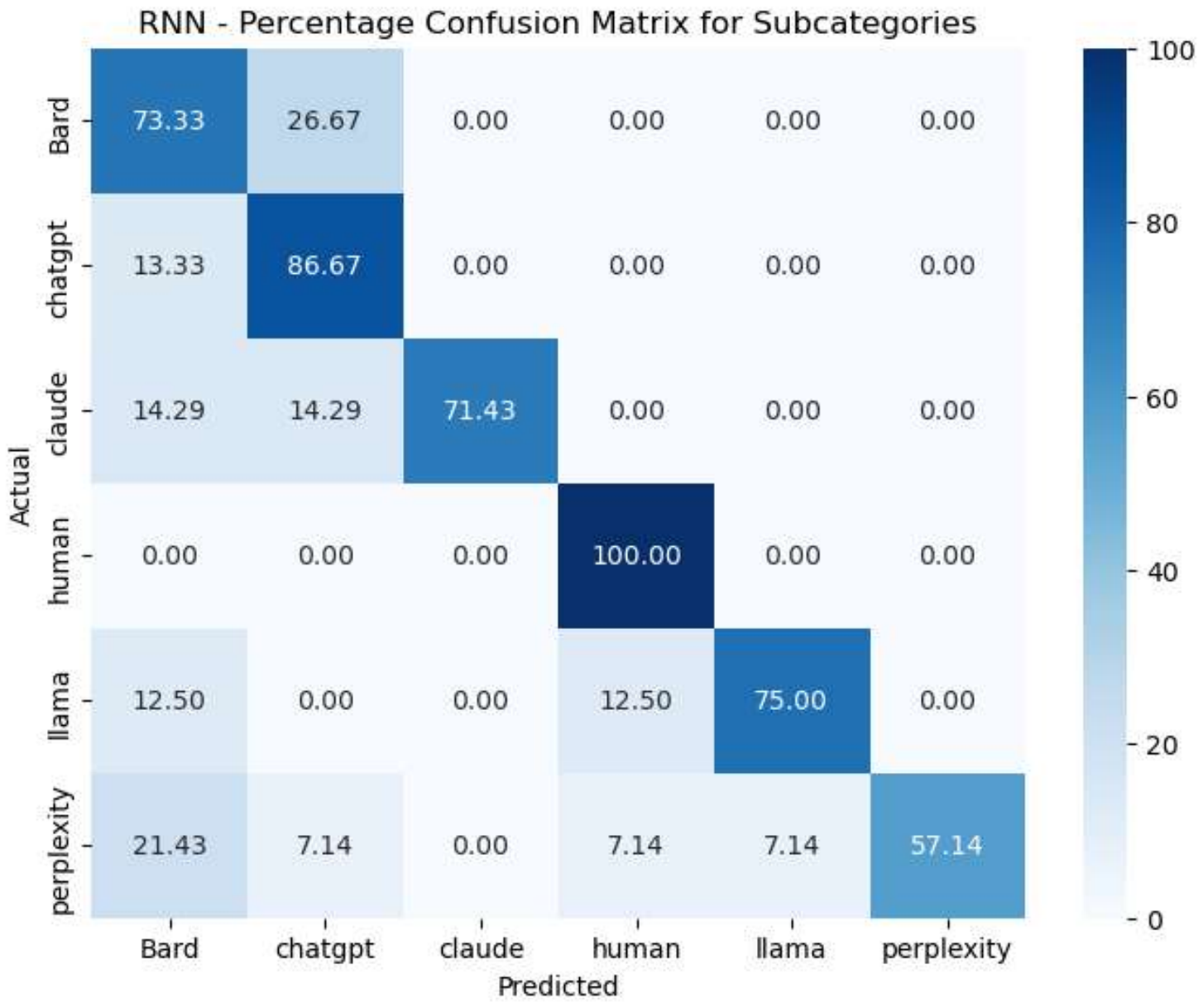}}  
\end{center}
\caption{Confusion matrices for multi-classification in percentage for (a) RF, (b) XGBoost, and (c) RNN }
\label{F7}
\end{figure}

As shown in Table~\ref{tab3} and Figure~\ref{F7}, the multi-classification study aims to differentiate texts produced by five distinct LLMs: ChatGPT, LLaMA, Google Bard, Claude, and Perplexity. We give the ROC curve and conclusions from the confusion matrix. Notably, we saw an excellent TP rate for the RNN and an excellent TP rate for the RF and XGBoost algorithms. For the "human" class, all three algorithms showed an exact 100\% TP rate; for the other classes, they produced positive results; however, RNN showed a 12.5\% TP rate for "claude," suggesting that it confused "human," "chatgpt," and "Bard." Furthermore, a 62.5\% TP rate for “llama” indicated that it was difficult to differentiate it from the terms "human," "chatgpt", and "perplexity." These observations point out strong points and possible misunderstandings in the classification outcomes.


\begin{table}[]
\caption{Binary classification accuracy results}
\label{tab2}
\begin{tabular}{lllll}
\hline
Algorithm & Accuracy & Precision & Recall & F1-Score \\ \hline
RF        & 97\%    & 96\%     & 95\%  & 96\%    \\ \hline
XGBoost   & 98\%    & 97\%     & 98\%  & 98\%    \\ \hline
RNN       & 94\%    & 93\%     & 94\%  & 94\%    \\ \hline
\end{tabular}
\end{table}


\begin{table}[]
\caption{Multi-classification accuracy results}
\label{tab3}
\begin{tabular}{lllll}
\hline
Algorithm & Accuracy & Precision & Recall & F1-Score \\ \hline
RF        & 97\%     & 93\%      & 94\%   & 93\%     \\ \hline
XGBoost   & 94\%     & 90\%      & 90\%   & 89\%     \\ \hline
RNN       & 88\%     & 90\%      & 72\%   & 74\%     \\ \hline
\end{tabular}
\end{table}


\subsection{Explainable AI}

In this part we examined how the RF algorithm used in this study affects the multiclassification detection, we will employ LIME for XAI in our investigation. By providing insights into the reasons impacting the model's conclusions in the field of plagiarism detection, this technique improves transparency. 

\begin{figure}[t]
\begin{center}
\subfloat[]{\includegraphics[width = 1.8 in]{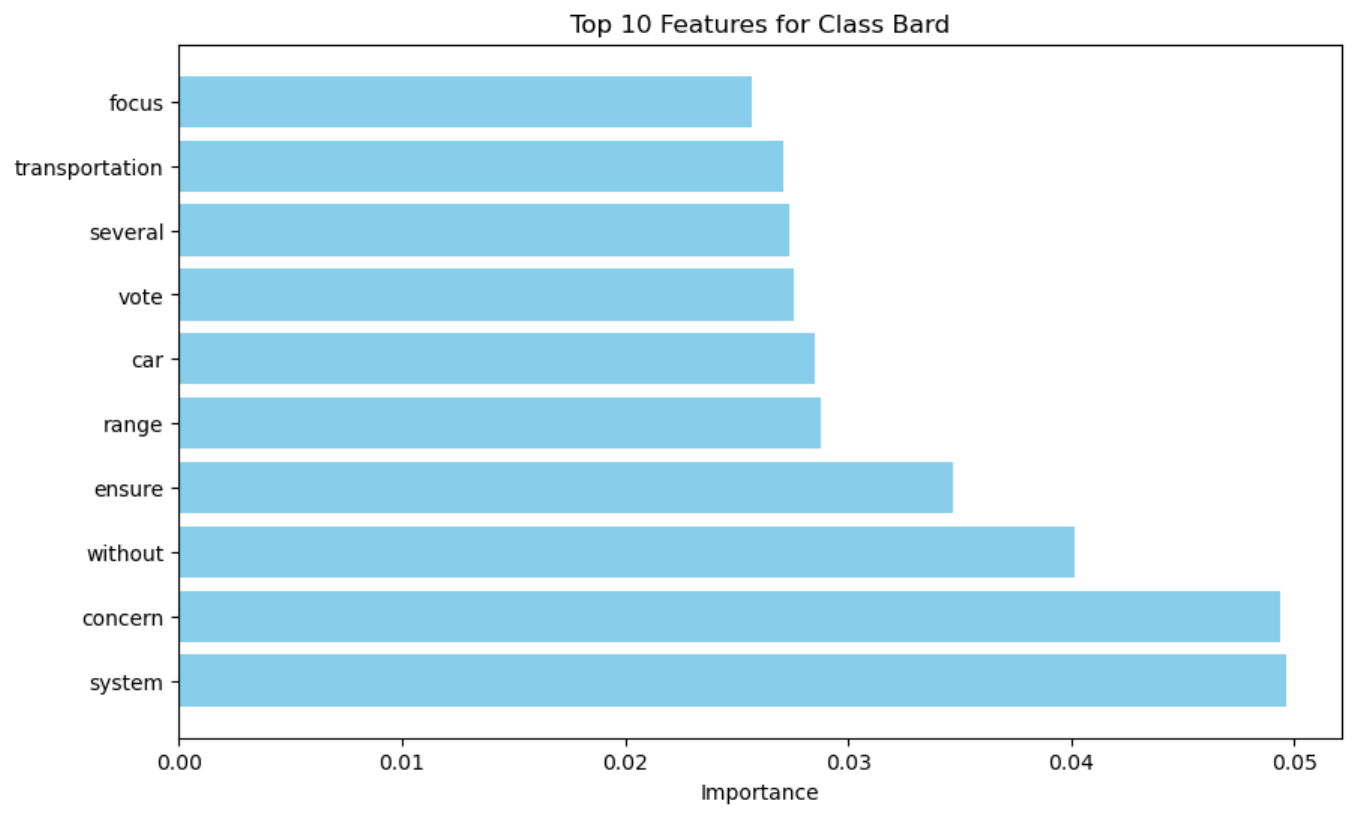}}
\subfloat[]{\includegraphics[width = 1.8 in]{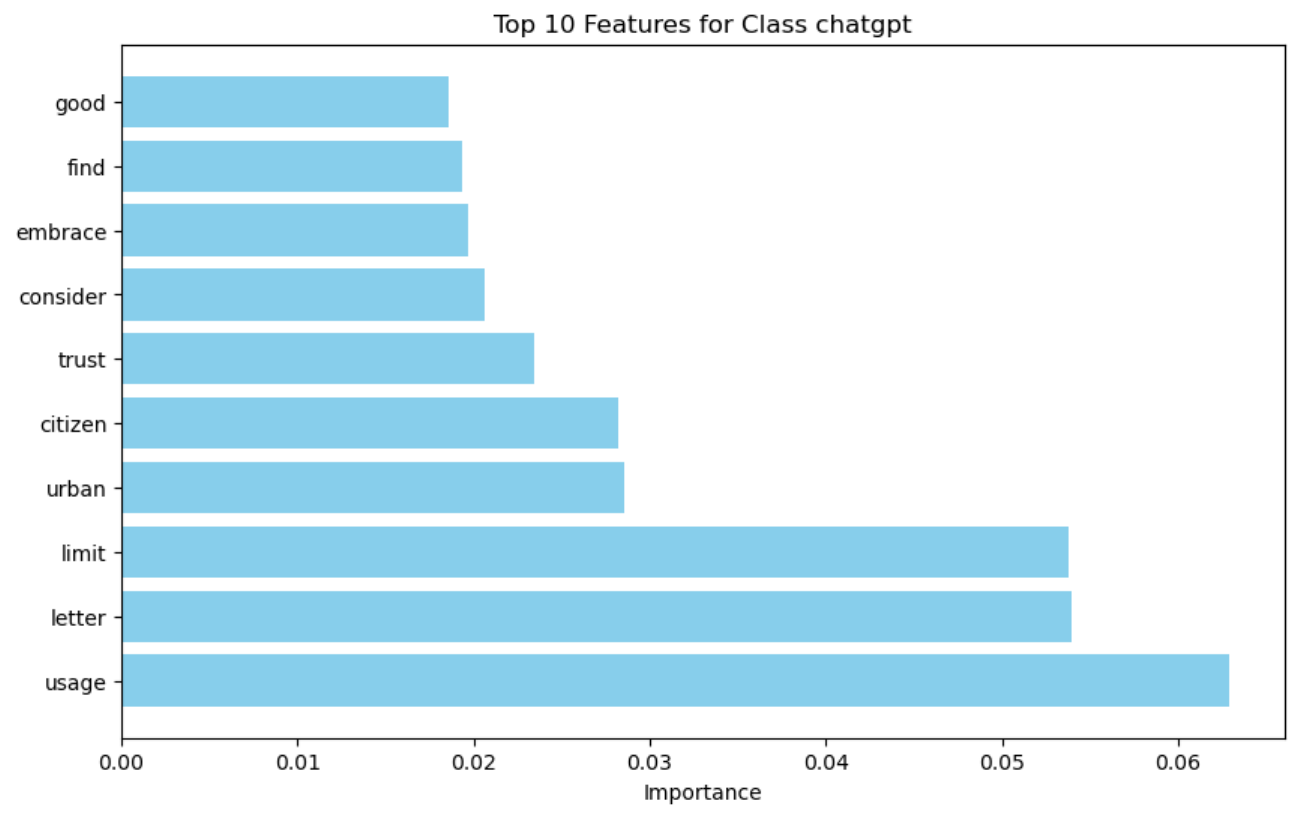}} 

\medskip
\subfloat[]{\includegraphics[width = 1.8 in]{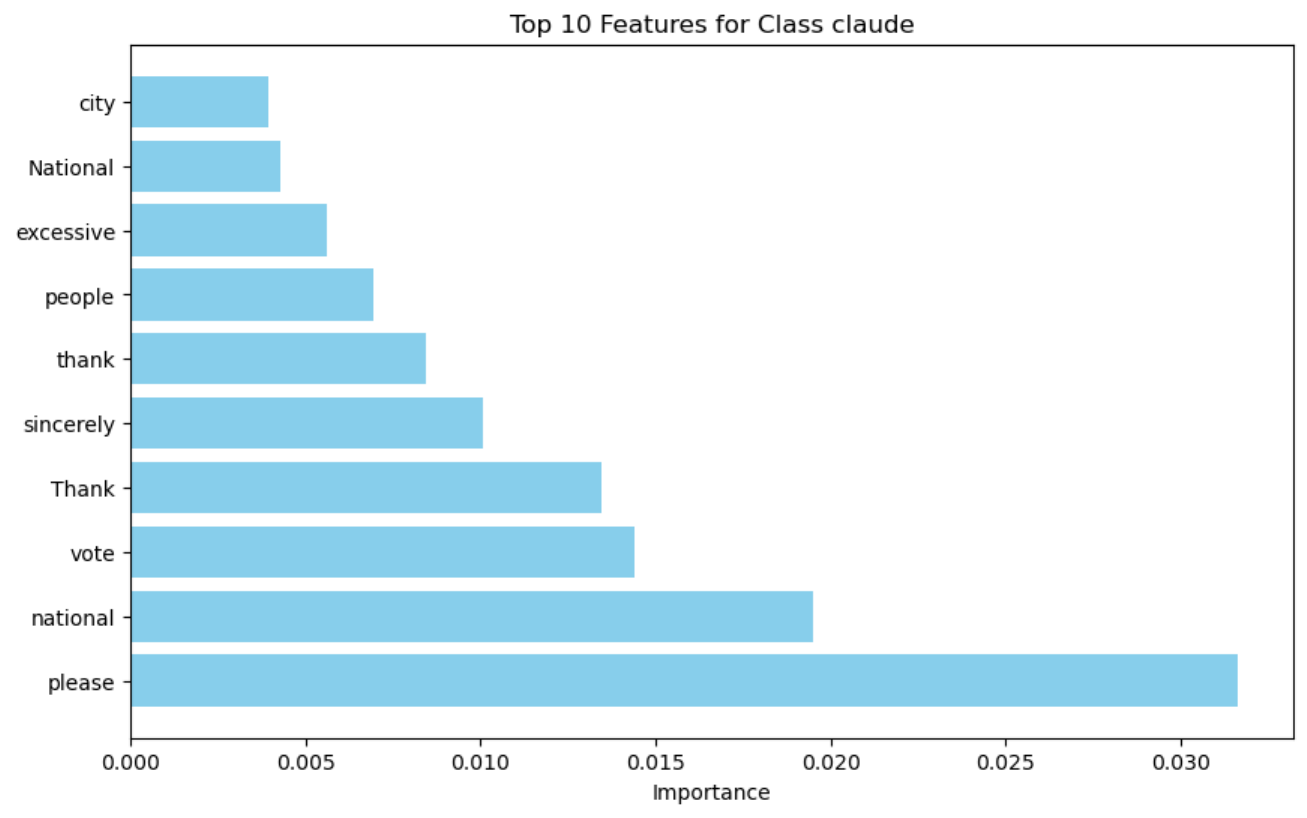}}  
\subfloat[]{\includegraphics[width = 1.8 in]{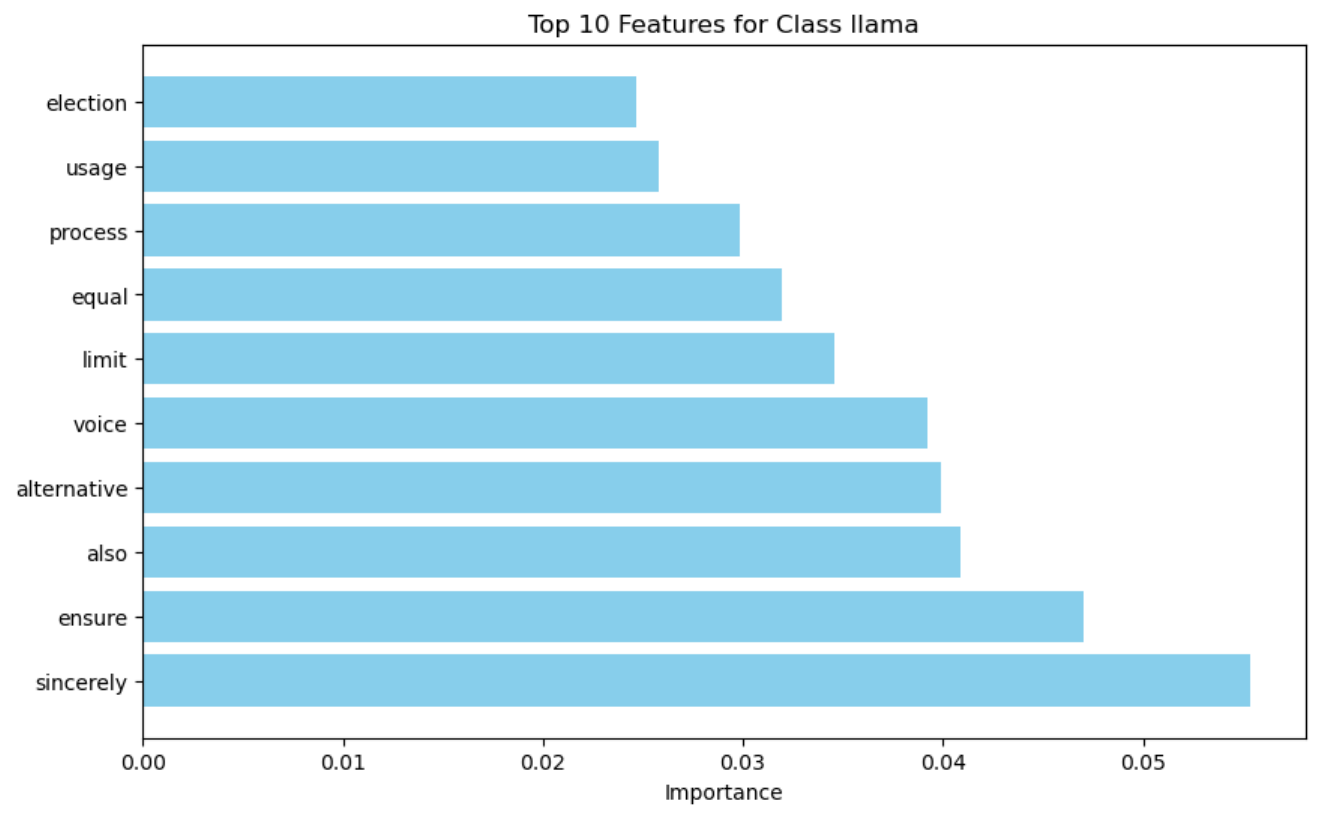}}

\medskip
\subfloat[]{\includegraphics[width = 1.8 in]{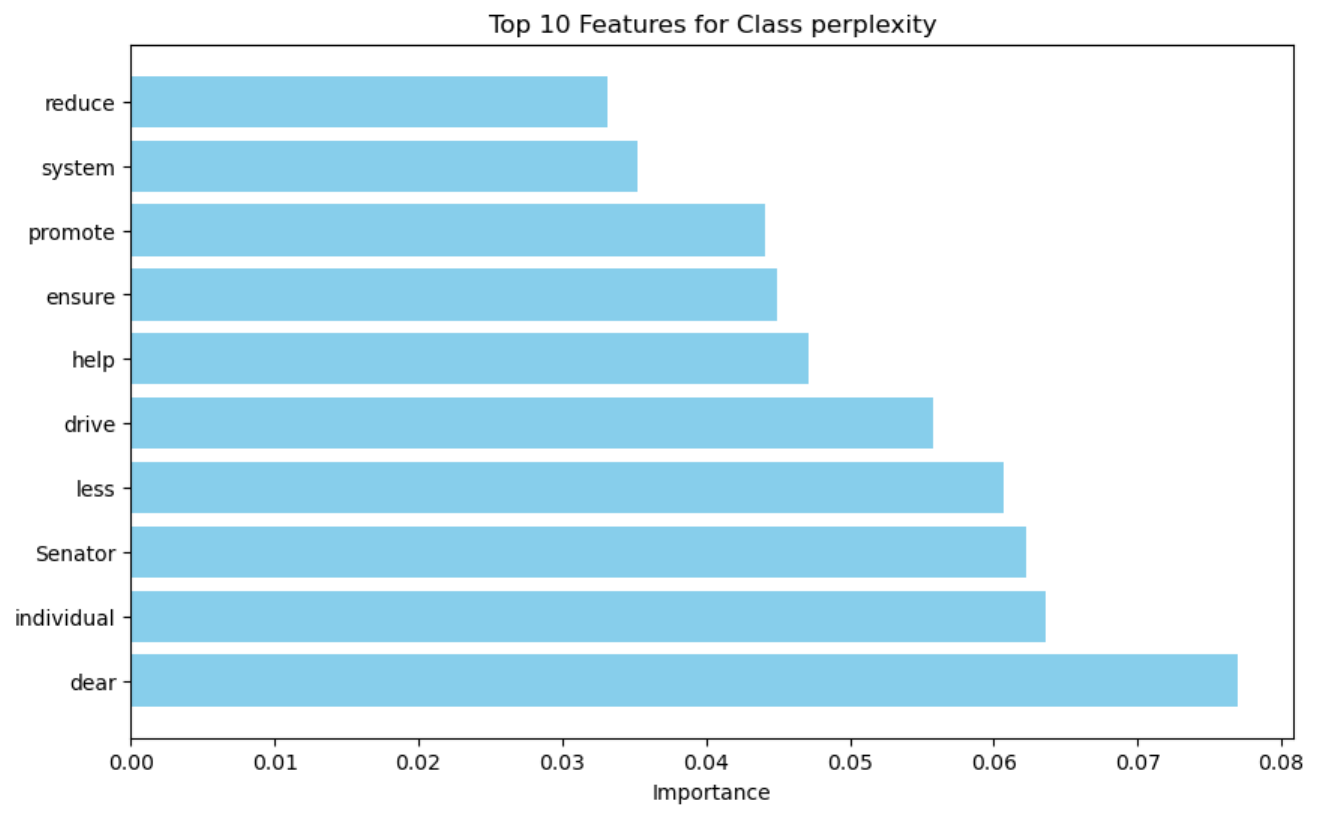}} 
\subfloat[]{\includegraphics[width = 1.8 in]{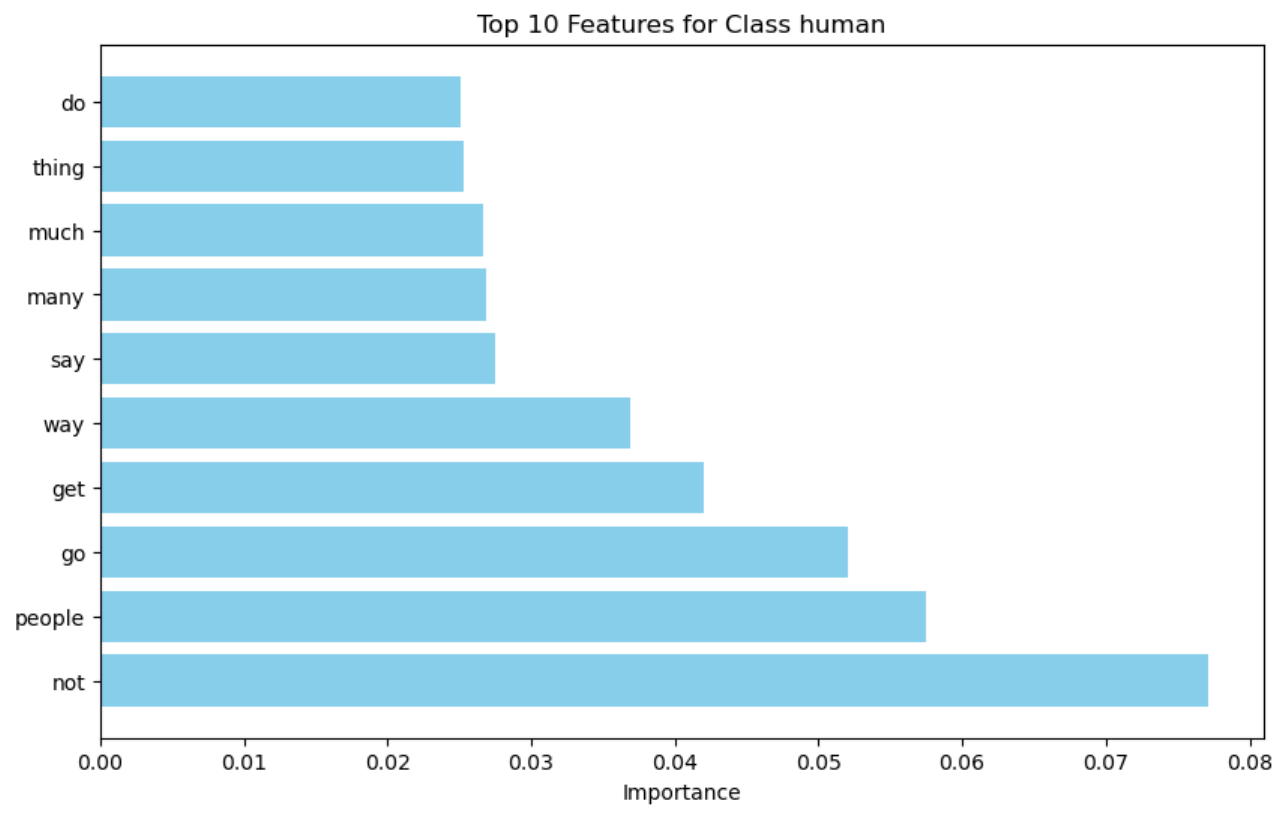}}  
\end{center}
\caption{The top 10 important features for the (a)"Bard", (b)”chatgpt”, (c) claude, (d)”llama”, (e) “perplexity”, and (f) “human” classes in the RF model  }
\label{F9}
\end{figure}

Figure~\ref{F9} shows the top 10 important features for the "chatgpt", "Bard" , "claude", "llama", "preplexity", and "human" classes in the RF model, as generated by LIME. These visualizations clarify the precise words that have a major influence on the model's predictions in each class using bar plots. These visualizations, which take advantage of LIME's local interpretability, help to clarify the decision-making process of the black-box model by providing succinct insights into the critical characteristics driving classification results for various text categories. 


The bar charts offer insights into the most critical features for predicting each class in RF model, highlighting the unique and shared features that are important across the different classes of "Bard," "chatgpt," "claude," "llama," "perplexity," and "human." These insights reveal significant differences and similarities in feature importance, reflecting the distinct characteristics and commonalities among the classes. The top features for "Bard" include terms related to systematic and structural elements such as "focus," "transportation," "several," "vote," "car," "range," "ensure," "without," "concern," and "system." The emphasis on "system" suggests a focus on organized, methodological aspects. For "chatgpt," the critical features include "good," "find," "embrace," "consider," "trust," "citizen," "urban," "limit," "letter," and "usage." These features highlight a mixture of qualitative assessments (e.g., "good," "trust") and operational terms (e.g., "usage," "urban"), suggesting a balance between subjective evaluation and practical application in the "chatgpt" context. The important features for "claude" include words like "city," "National," "excessive," "people," "thank," "sincerely," "Thank," "vote," "rational," and "please." This class appears to prioritize social and formal interaction elements, with a notable emphasis on politeness and civic engagement (e.g., "please," "vote," "thank"). "Llama" is characterized by features such as "election," "usage," "process," "equal," "limit," "voice," "alternative," "also," "ensure," and "sincerely." These features suggest a focus on procedural and democratic elements, reflecting themes of equality, participation, and alternative approaches. The top features for "perplexity" include "reduce," "system," "pressure," "ensure," "help," "drive," "less," "Senator," "individual," and "dear." This class seems to emphasize reduction and efficiency (e.g., "reduce," "less"), systemic approaches, and individual importance, indicating a concern with optimization and personal significance. For "human," the critical features include "do," "thing," "much," "many," "say," "way," "get," "go," "people," and "not." These features reflect a focus on action and communication, with a high frequency of common verbs and pronouns, indicating practical, everyday human activities and interactions. 

However, across all classes, certain themes such as procedural terms ("ensure," "system") and evaluative terms ("consider," "trust," "good") appear frequently. Words like "ensure" and "usage" are common in multiple classes, indicating their broad relevance and importance in predicting various outcomes. The "Bard" and "claude" classes have unique terms that focus on structure and civility, respectively, highlighting their distinct contexts. "Chatgpt" and "llama" include more operational and procedural terms, reflecting their applications in practical and democratic settings. "Perplexity" and "human" classes emphasize efficiency and everyday human activities, respectively, indicating their specific focuses on optimization and practical action. 

\begin{figure}[t]
\begin{center}
\subfloat[]{\includegraphics[width = 1.8 in]{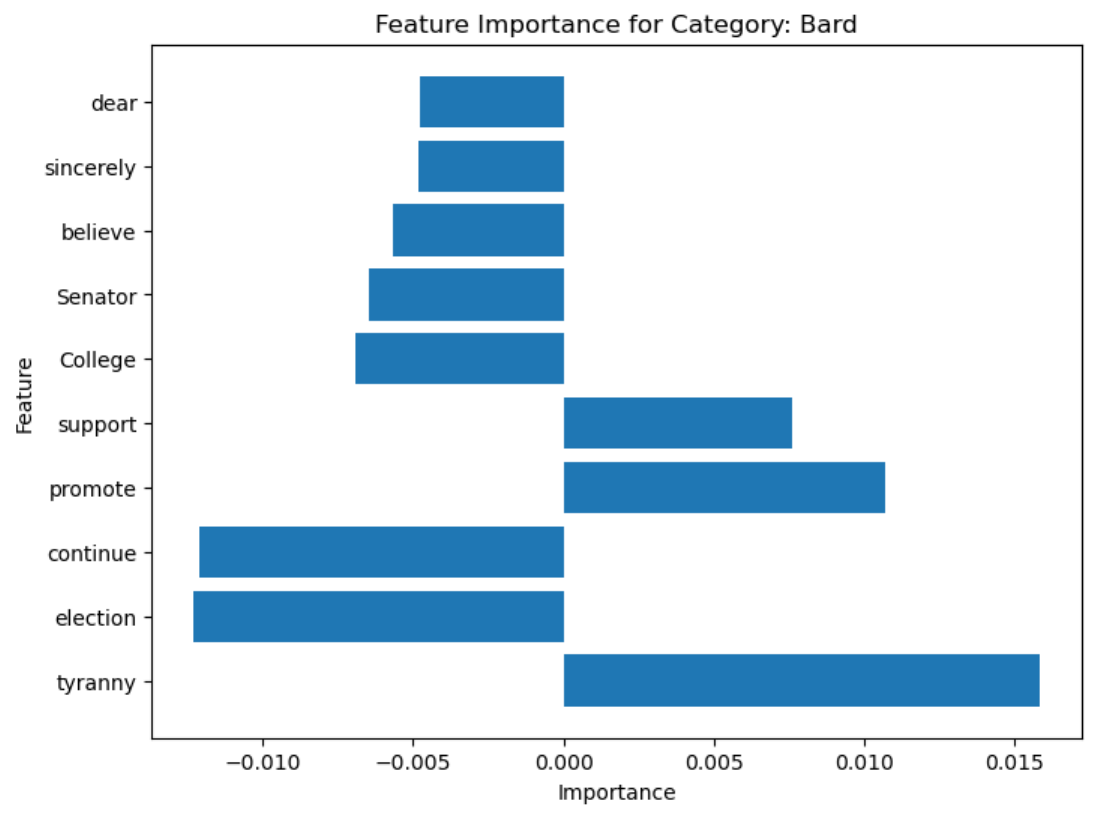}}
\subfloat[]{\includegraphics[width = 1.8 in]{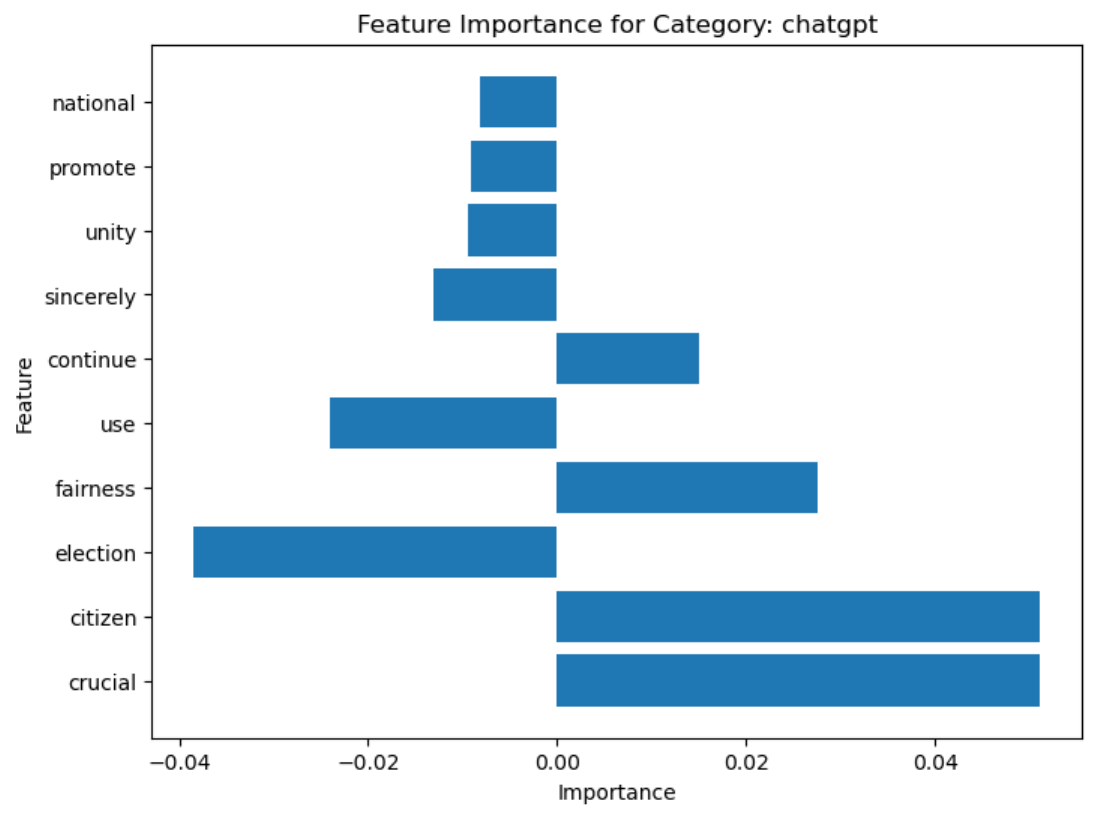}} 

\medskip
\subfloat[]{\includegraphics[width = 1.8 in]{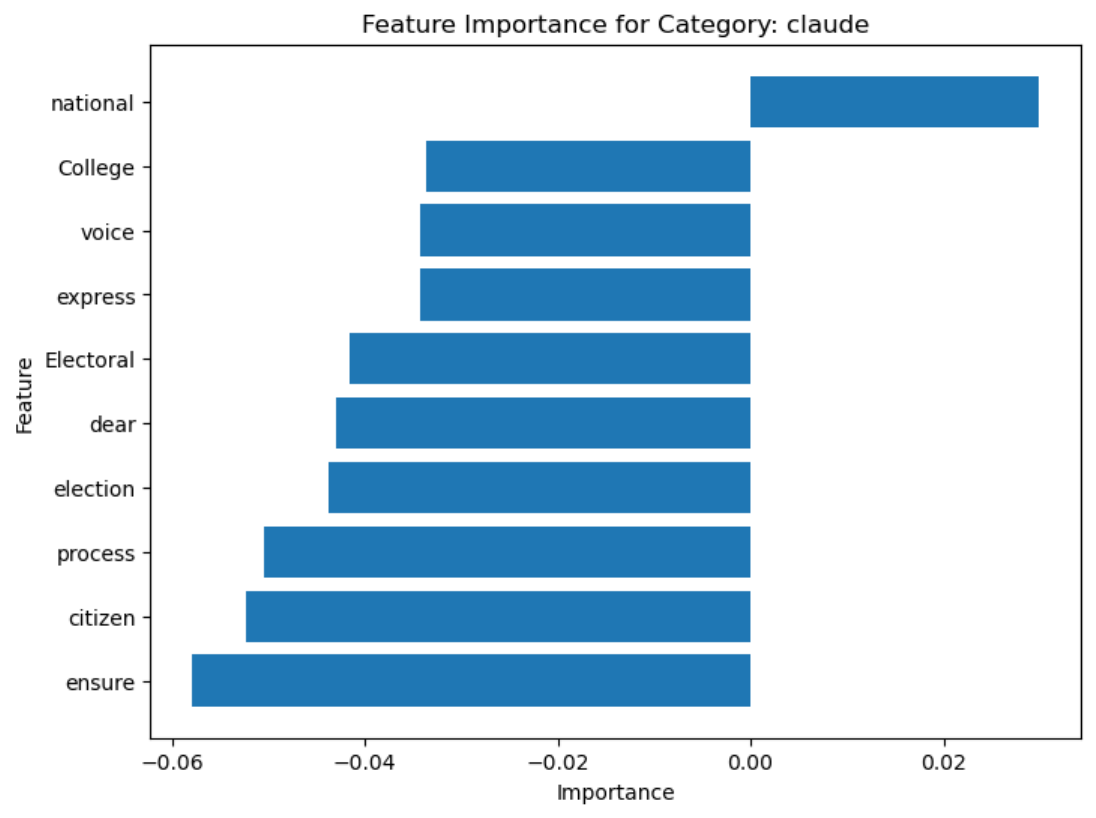}}  
\subfloat[]{\includegraphics[width = 1.8 in]{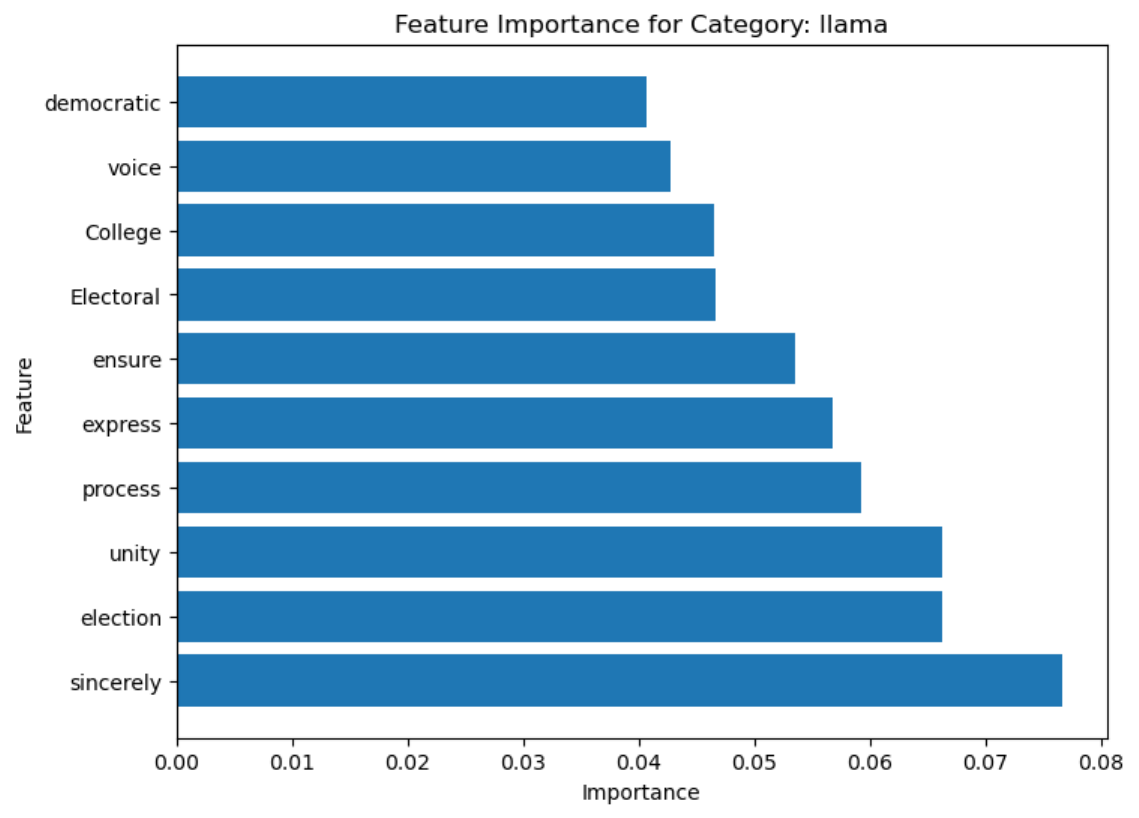}}

\medskip
\subfloat[]{\includegraphics[width = 1.8 in]{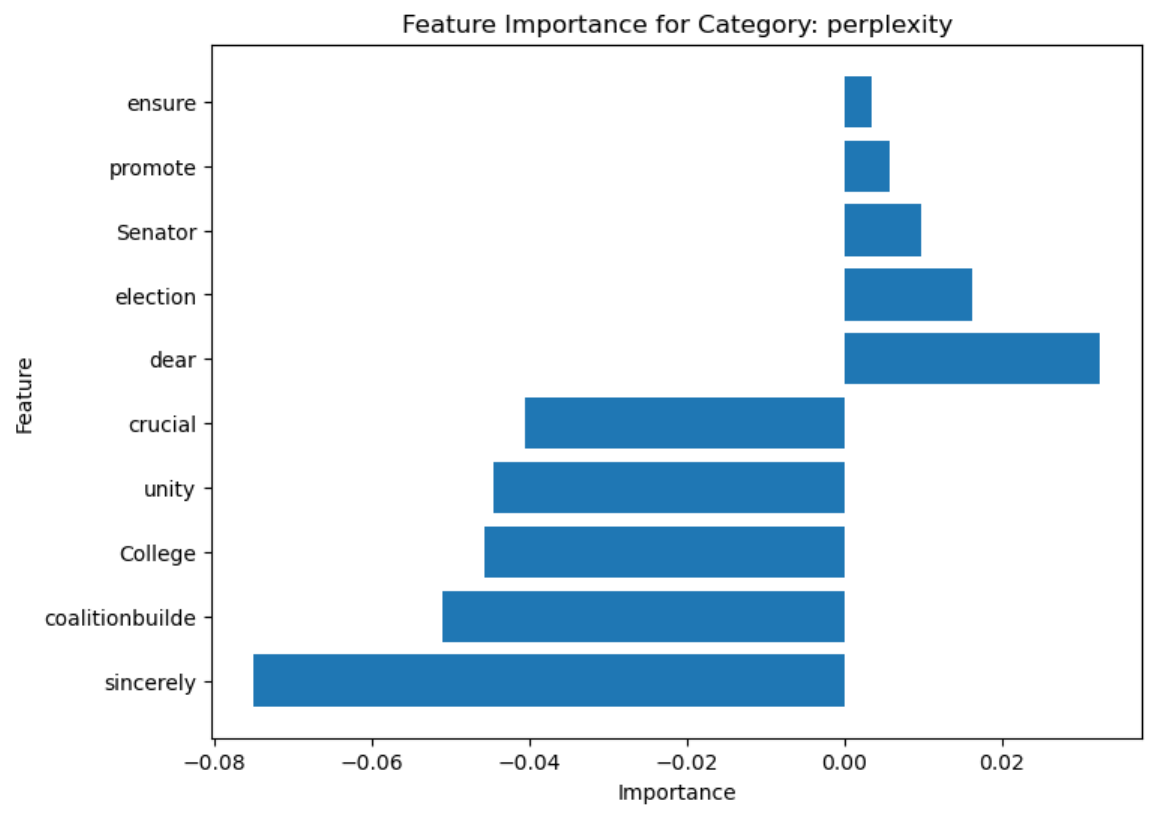}} 
\subfloat[]{\includegraphics[width = 1.8 in]{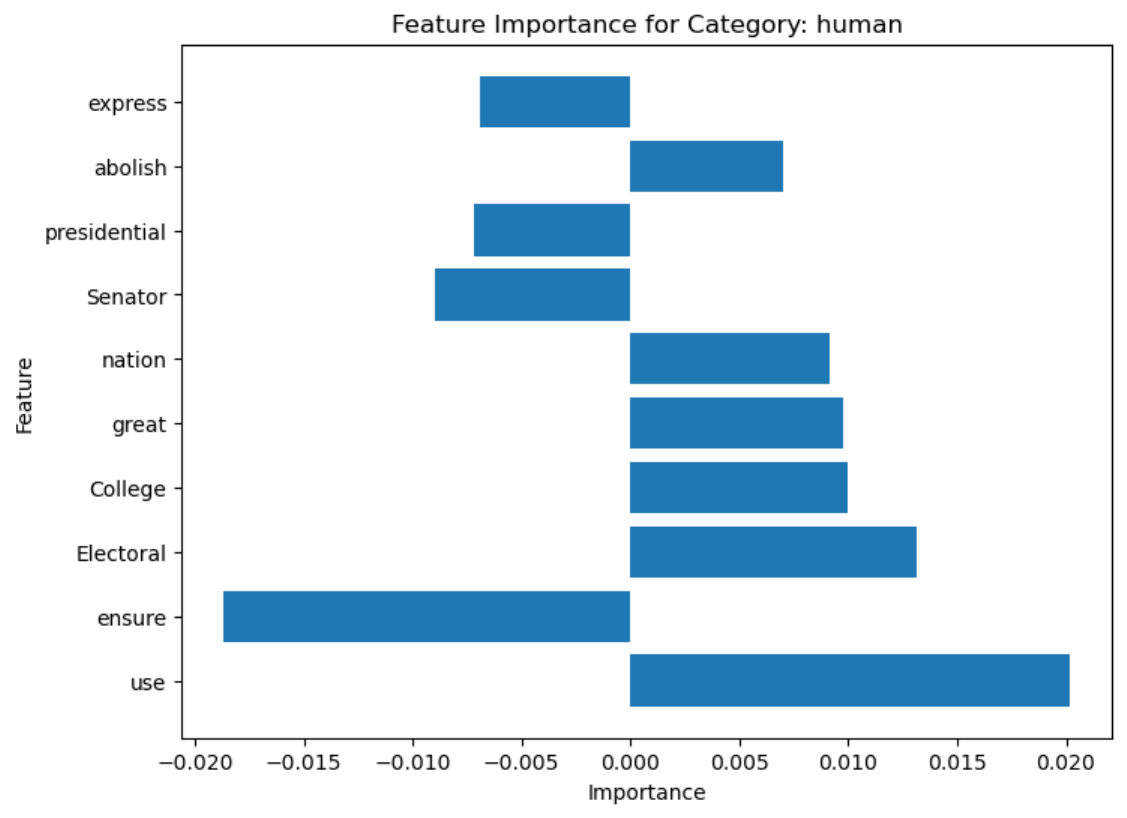}}  
\end{center}
\caption{The top 10 important features for the (a)"Bard", (b)”chatgpt”, (c) claude, (d)”llama”, (e) “perplexity”, and (f) “human” classes in the RF model for a specific instance.}
\label{F10}
\end{figure}

Table~\ref{tab7} and Figure~\ref{F10} explain the selected text instance using LIME. The output provides insight into the local decision-making process of the model and consists of the instance text, true label, predicted label, and a visual representation of the top ten features impacting the prediction respectively. This graphic, which has its basis in LIME explanations, aids in the interpretation of the data by making clear the specific terms that significantly influence the expected result. 


\begin{table}[]
\caption{ The prediction probabilities for a specific instance}
\label{tab7}
\resizebox{%
      \ifdim\width>\columnwidth
        \columnwidth
      \else
        \width
      \fi
    }{!}{%
\begin{tabular}{|l|l|l|l|}
\hline
Instance text & True label & Predicted label & Prediction probabilities \\ \hline
\begin{minipage}[t]{0.6\columnwidth}%
“dear   Senator write express strong support continue use Electoral College   presidential election process proud citizen great nation believe crucial   maintain principle fairness equality upon democracy found Electoral College   ensure small state voice election process promote coalitionbuilde national unity serve vital check tyranny majority implore uphold integrity democratic system reject attempt abolish Electoral College sincerely” %
\end{minipage} & llama  & llama & 
\includegraphics[width = 1.8 in]{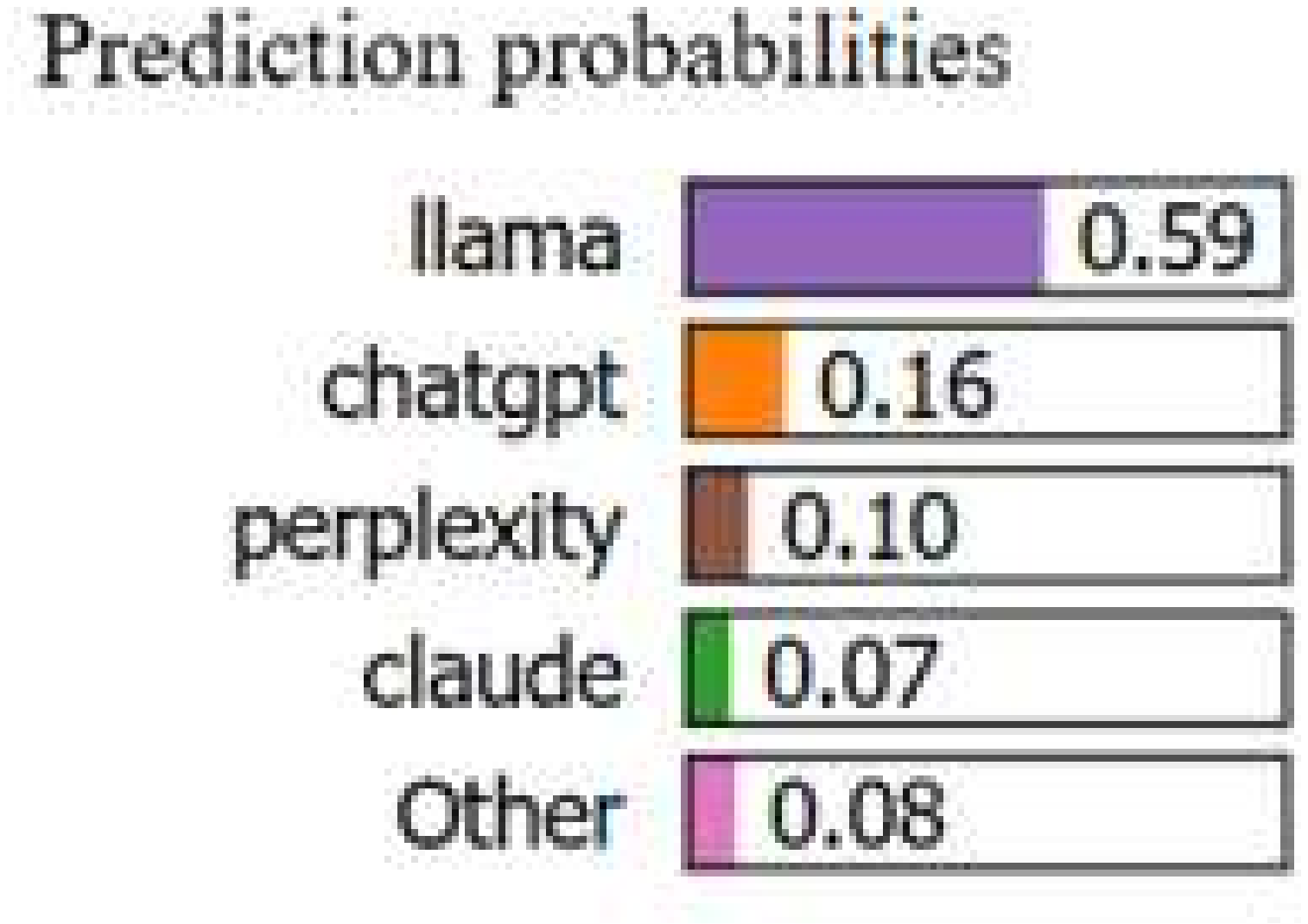}

\\ \hline                  
\end{tabular}}
\end{table}


The analysis of feature importance across different classes in the RF model can be highly practical for detecting plagiarism. By understanding the key features that distinguish various classes, we can develop robust models to identify and flag potentially plagiarized content. Each chart provides insights into the most critical features for predicting each class, such as "ensure," "system," and "trust" for Bard, or "good," "find," and "embrace" for ChatGPT. These terms might be consistently used by specific LLM tool, making them important markers of their writing style.  

By analyzing the most important features for different classes, we can create detailed profiles of each LLM. These profiles can help in identifying stylistic and structural elements unique to specific sources. When new content is analyzed, the model can compare it against these established profiles to detect discrepancies or similarities that might indicate plagiarism. This method not only highlights direct copying but also more subtle forms of plagiarism where the content has been paraphrased or lightly edited. Thus, the detailed understanding of feature importance aids in creating a nuanced and effective plagiarism detection system. 

\subsection{Evaluation }

One of the most challenging duties was addressing in this section: comparing our model's accuracy to standards set by the industry. To exceed their accuracy levels, we benchmarked against the GPTZero~\cite{gptzero, GPTZero2}, in our comparison. We are certain that our prototype has the potential to be further developed and scalable for commercial use, even with a smaller training dataset than industry-level norms.

GPTZero is an application of AI detection software that Edward Tian, an undergraduate student at Princeton University, created to identify artificially generated text, especially from huge language models. GPTZero, which was introduced in January 2023 to address worries about AI-driven academic plagiarism, has received praise for its work but has also drawn criticism for producing false positives, particularly in situations where academic integrity is at risk. The program uses burstiness and perplexity metrics to identify passages that are created by bots~\cite{GPTZerow}. Burstiness examines phrase patterns for differences, whereas Perplexity measures text randomization and odd construction based on language model prevalence. Human text has greater diversity than content generated by AI. In previous studies comparing GPTZero and ChatGPT's efficacy in assessing fake queries and medical articles~\cite{habibzadeh2023gptzero}, GPTZero was utilized in multiple investigations. This study found that GPTZero had low false-positive and high false-negative rates. A second analysis of more than a million tweets and academic papers looked at opinions regarding ChatGPT's capacity for plagiarism~\cite{heumann2023chatgpt}. It contrastedit with the lack of interest in GPTZero, intended to prevent plagiarism caused by artificial intelligence. The difficulties and possibilities for both models were determined using sophisticated natural language processing techniques, providing information for further conversational AI research. The used classes in GPTZero can be seen in Table~\ref{tab5}.

In the binary classification, we used GPTZero to recognize text. To do this, a test dataset that makes up 20\% of the original dataset used for binary classification must be extracted. A summary of the detection process's results may be seen in Table~\ref{tab6}. While GPTZero showed about 78.3\% accuracy not identifying every occurrence, our algorithms showed a higher accuracy of about 97.5\%.


\begin{table*}[]
\caption{GPTZero classes}
\label{tab5}
\resizebox{\textwidth}{!}{%
\begin{tabular}{lll} \hline
Class Name       & GPTZero Message                                                                                                            & AI text percentage                         \\ \hline
Human            & “This text is most likely to be written by a human”                                                                        & 0-10\%                                     \\ \hline
Different Result & “Our ensemble of detectors predicts different results for this text. Please enter more text for more precise predictions.” & 11-39\%                                    \\ \hline
Mix              & “This text is likely to be a mix of human and AI text”                                                                     & 40-88\%                                    \\ \hline
AI               & “This text is likely to be written by AI”                                                                                  & 89-100\%                                   \\ \hline
Not Recognized   & "Try typing in some more text (\textgreater{}250 characters) so we can give you accurate results"                          & The total text is less than 250 characters \\ \hline
\end{tabular}
}
\end{table*}


\begin{table}[]
\caption{Comparison with the GPTZero tool}
\label{tab6}
\resizebox{%
      \ifdim\width>\columnwidth
        \columnwidth
      \else
        \width
      \fi
    }{!}{%
\begin{tabular}{|l|l|l|l|l|l|l|l|}
\hline
                           & Class & Human & AI & Mix & Different Result & Not Recognized & Accuracy                \\ \hline
\multirow{2}{*}{GPTZero}   & Human & 59    & 1  & 0   & 8                & 0              & \multirow{2}{*}{78.3\%} \\ \cline{2-7}
                           & LLMs  & 5     & 35 & 7   & 0                & 5              &                         \\ \hline
\multirow{2}{*}{Our Model} & Human & 66    & 2  & -   & -                & -              & \multirow{2}{*}{97.5\%}  \\ \cline{2-7}
                           & LLMs  & 1     & 51 & -   & -                & -              &                         \\ \hline
\end{tabular}
}
\end{table}
 
\section{Conclusion}

In this study, we aimed to address the growing challenge of textual content attribution in a world where LLMs are increasingly used to generate content. We compared ML and DL algorithms demonstrating how a fine-tuned, targeted approach can outperform more generalized models in accurately classifying text and ensuring transparency in the decision-making process through XAI techniques. This is achieved by looking into how to identify these texts using ML and XAI techniques in two sections: one for multi-classification, where we differentiate between five different LLM tools (ChatGPT, LLaMA, Google Bard, Claude, and Perplexity) and human-written text we found that the RF gives the best result with 97\% accuracy, and another for binary classification, where we distinguished between text generated by LLMs generally and text written by humans and the three algorithms RF, XGBoost and RNN achieving a highest accuracy of 98\% accuracy. Our model outperformed GPTZero with 98.5\% accuracy to 78.3\%. Notably, GPTZero was unable to recognize about 4.2\% of the observations, but our model was able to recognize the complete test dataset.

XAI showed that there are critical features for predicting each class, such as "ensure," "system," and "trust" for Bard, or "good," "find," and "embrace" for ChatGPT. These terms might be consistently used by specific LLM tool, making them important markers of their writing style. Analyzing feature importance across various classes provides valuable insights for detecting plagiarism by identifying distinct stylistic and structural elements unique to different authors. This approach not only helps in solving the textual content attribution and spotting direct copying but also in recognizing more subtle forms of plagiarism, thereby ensuring thorough and accurate verification of content originality.



\bibliographystyle{apacite}
 \bibliography{cas-refs}





\end{document}